\documentclass[10pt,twocolumn,letterpaper]{article}

\usepackage{cvpr}              % To produce the CAMERA-READY version
\newlength\savewidth\newcommand\shline{\noalign{\global\savewidth\arrayrulewidth
  \global\arrayrulewidth 1pt}\hline\noalign{\global\arrayrulewidth\savewidth}}
\newcommand{\tablestyle}[2]{\setlength{\tabcolsep}{#1}\renewcommand{\arraystretch}{#2}\centering\footnotesize}
\renewcommand{\paragraph}[1]{\noindent\textbf{#1}}

\newcommand{\nbf}[1]{{\noindent \textbf{#1}}}

\definecolor{cvprblue}{rgb}{0.21,0.49,0.74}
\usepackage[pagebackref,breaklinks,colorlinks,allcolors=cvprblue]{hyperref}

\def\paperID{4540} % *** Enter the Paper ID here
\def\confName{CVPR}
\def\confYear{2025}

\title{GIVEPose: Gradual Intra-class Variation Elimination for RGB-based Category-Level Object Pose Estimation}

\author{
Ziqin Huang$^{1}$ \qquad
Gu Wang$^{1}$ \qquad
Chenyangguang Zhang$^{1}$ \qquad
Ruida Zhang$^{1}$ \qquad
\\
Xiu Li$^{1,2}$ \qquad
Xiangyang Ji$^{1}$\\
\small
$^{1}$Tsinghua University \quad
$^{2}$Pengcheng Laboratory\quad \\
{\tt\small 
\{huang-zq24@mails., xyji@\}tsinghua.edu.cn}
}
\begin{document}
\maketitle
\begin{abstract}
Recent advances in RGBD-based category-level object pose estimation have been limited by their reliance on precise depth information, restricting their broader applicability. In response, RGB-based methods have been developed. Among these methods, geometry-guided pose regression that originated from instance-level tasks has demonstrated strong performance. However, we argue that the NOCS map is an inadequate intermediate representation for geometry-guided pose regression method, as its many-to-one correspondence with category-level pose introduces redundant instance-specific information, resulting in suboptimal results. This paper identifies the intra-class variation problem inherent in pose regression based solely on the NOCS map and proposes the Intra-class Variation-Free Consensus (IVFC) map, a novel coordinate representation generated from the category-level consensus model. By leveraging the complementary strengths of the NOCS map and the IVFC map, we introduce GIVEPose, a framework that implements Gradual Intra-class Variation Elimination for category-level object pose estimation.
Extensive evaluations on both synthetic and real-world datasets demonstrate that GIVEPose significantly outperforms existing state-of-the-art RGB-based approaches, achieving substantial improvements in category-level object pose estimation.
Our code is available at \url{https://github.com/ziqin-h/GIVEPose}.
\end{abstract}    
\section{Introduction}
\label{sec:intro}

The estimation of object pose is a critical task in computer vision, with a wide range of applications in robotic manipulation~\cite{mousavian20196,wen2022you,zhai2023monograspnet,fu_iros22_assembly,fu_eccvw24_LanPose}, augmented reality~\cite{marchand2015pose,tang20193d}, and autonomous driving~\cite{manhardt2019roi,wu20196d}. 
Over the past few years, numerous deep learning-based approaches have been developed to tackle this task. 
Initial research~\cite{xiang2017posecnn, rad2017bb8, peng2019pvnet, li2019cdpn, park2019pix2pose, wang2019densefusion, wang2021gdr} primarily focused on instance-level tasks, yielding impressive results. 
However, this setting faced significant limitations due to its reliance on CAD models and its inability to generalize to previously unseen objects. 
Consequently, category-level object pose estimation has been proposed to estimate the poses of unseen objects within predefined categories.

\begin{figure}[t]
  \centering
   \includegraphics[width=1.0\linewidth]{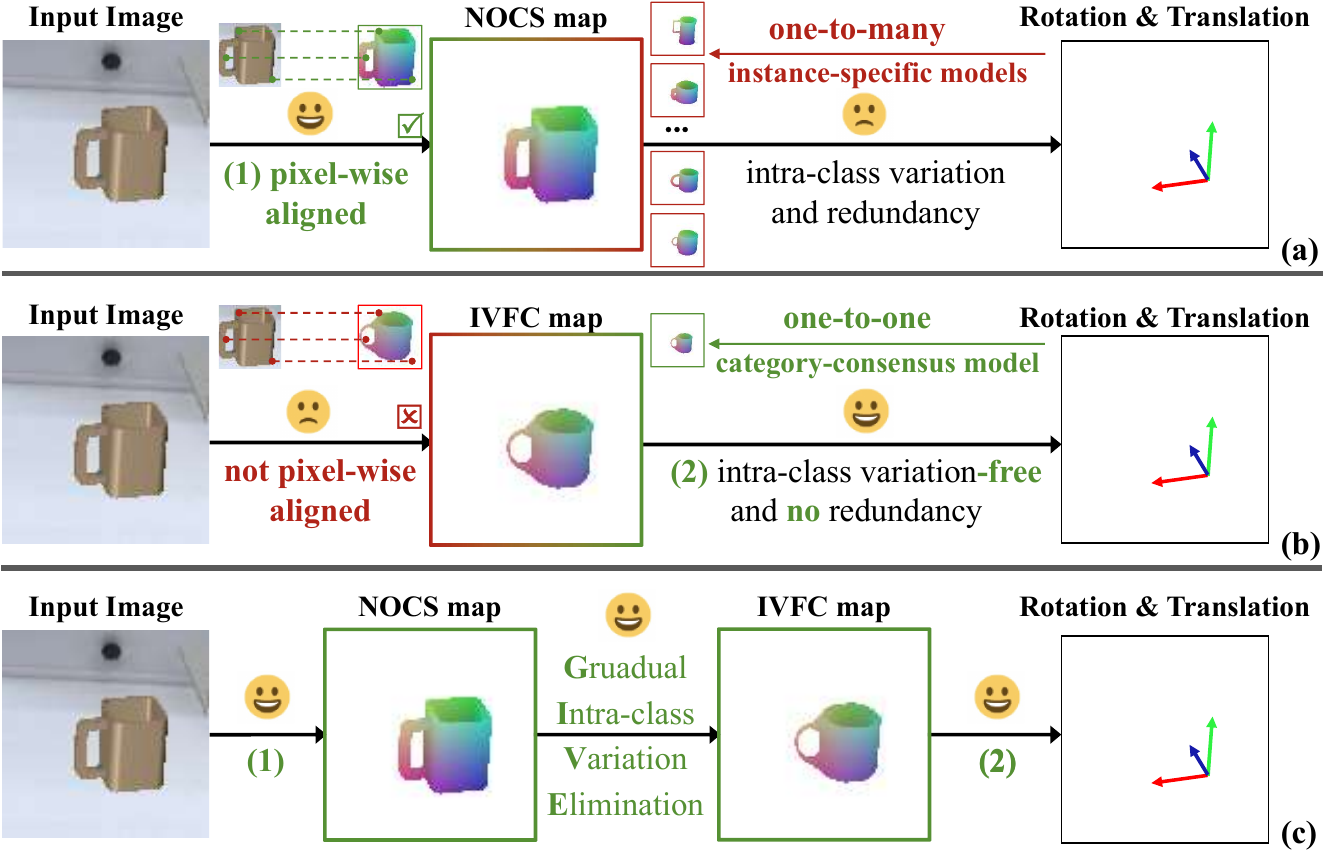}

   \caption{Comparison of different coordinate representations as intermediate supervision. (a) Category-level pose prediction from the NOCS map alone suffers from intra-class variation and redundancy effects. (b) The prediction of intermediate coordinate maps becomes intractable when utilizing IVFC map exclusively, due to its lack of pixel-wise alignment with input images. (c) Our proposed gradual intra-class variation elimination strategy leverages the complementary advantages of both maps, enabling more precise category-level pose estimation.
   }
   \label{fig:tri}
\end{figure}

Wang et al.~\cite{wang2019normalized} pioneered category-level pose estimation with the introduction of Normalized Object Coordinate Space (NOCS) and proposed benchmark datasets for validation. 
Subsequently, numerous RGBD-based category-level pose estimation methods \cite{lin2023vi,liu2023ist,lin2024ag, chen2024secondpose} have demonstrated remarkable performance. 
However, these approaches are constrained by their reliance on depth information, limiting their broader applicability. 
To address this limitation, several recent studies \cite{lee2021category, fan2022object, wei2023rgb, zhang2024lapose} have explored category-level pose estimation solely using RGB images.
\cite{lee2021category, fan2022object, wei2023rgb} first establish dense 2D-3D or 3D-3D correspondences and subsequently determine object pose using appropriate algorithms~\cite{epnp2009ep,umeyama1991least}. 
In contrast to these indirect approaches, LaPose~\cite{zhang2024lapose} adapts the instance-level geometry-guided pose regression method for category-level tasks and achieves state-of-the-art performance.

In LaPose~\cite{zhang2024lapose}, the NOCS map is modeled using a Laplacian Mixture Model (LMM) to mitigate the effects of intra-class shape variation on NOCS map estimation and then the pose is predicted based on the LMM-modeled NOCS map. However, we argue that the intra-class variation of the NOCS map itself also impedes category-level pose estimation within a geometry-guided regression framework, making it an inadequate intermediate representation. As illustrated in \cref{fig:tri} (a) and \cref{fig:iv}, significant variance exists among the NOCS maps of different instances for the same rotation and translation due to intra-class shape variation. Consequently, predicting category-level rotation and translation solely through NOCS-map-based correspondences introduces redundant instance-specific information, which is detrimental to category-level pose estimation.

According to the analyses from cognitive psychology research~\cite{rosch1978principles}, humans recognize objects primarily based on coarse-grained category-shared information (i.e., prototypes~\cite{prototype1, prototype2}) rather than fine-grained features. Inspired by this, we argue that the NOCS map encompasses both fine-grained instance-specific information and coarse-grained category-consensus information whereas only the latter is relevant for estimating category-level pose. 
Therefore, we propose the \textbf{I}ntra-class \textbf{V}ariation-\textbf{F}ree \textbf{C}onsensus (IVFC) map, generated by the category-level consensus model, which maintains a one-to-one relationship with category-level pose while excluding redundant instance-specific information.
% as illustrated in \cref{fig:iv} (right). 

Despite the aforementioned benefits of the IVFC map, the lack of pixel-wise correspondence between this map and the input image presents significant challenges when utilizing the IVFC map exclusively as an intermediate representation for direct prediction from the cropped image, as shown in \cref{fig:tri}(b). 
To this end, we accordingly propose \textbf{GIVEPose}, a novel approach that employs \textbf{G}radual \textbf{I}ntra-class \textbf{V}ariation \textbf{E}limination to facilitate RGB-based category-level \textbf{Pose} estimation (see \cref{fig:tri}(c)). 
The overall network architecture is illustrated in \cref{fig:pipline}.
After obtaining cropped images of known category objects, we first feed them into the backbone network for feature extraction and subsequently estimate the NOCS map. 
Once we have the NOCS map, we introduce a \textbf{D}eformable \textbf{C}onvolutional \textbf{A}uto-\textbf{E}ncoder (DCAE)-based module to gradually eliminate instance-specific redundant information. 
Specifically, we employ a more flexible deformable convolution-based encoder to downsample the NOCS map, taking into account the non-pixel-wise nature of intra-class variation and the unique characteristics of the coordinate map. 
By fusing the encoder features with the backbone features, we utilize an upsampling network as decoder to reconstruct the IVFC map. 
The subsequent pose predictor integrates the IVFC map with the 2D positional information for the final category-level pose estimation.

To summarize, our main contributions are threefold:
\begin{itemize}
    \item We reveal the intra-class variation problem that arises when using the NOCS map as the sole intermediate representation for category-level pose regression and further propose the IVFC map to alleviate this issue.
    \item 
    Leveraging the complementary strengths of the NOCS map and the IVFC map, we introduce a novel approach termed GIVEPose and design a DCAE-based module for its implementation.
    \item Our experiments on the NOCS~\cite{wang2019normalized} and Wild6D~\cite{fu2022wild6d} datasets confirm our analysis of the properties of both coordinate maps, and our method achieves substantial improvements over state-of-the-art methods, showcasing its effectiveness.
\end{itemize}

\section{Related Work}

\subsection{Instance-level Object Pose Estimation}
Instance-level object pose estimation refers to estimating the 6D object pose from an image containing the object seen during training. 
For RGB image input, recent approaches can be roughly categorized into indirect two-stage pose estimation and regression-based pose estimation. 
The indirect two-stage approach first establishes either dense \cite{li2019cdpn, hodan2020epos,park2019pix2pose, haugaard2022surfemb,li2022egraph} or sparse \cite{rad2017bb8, peng2019pvnet, song2020hybridpose} 2D-3D correspondences and subsequently solves for pose using the PnP/RANSAC algorithm. 
Among the regression-based methods, some directly estimate pose \cite{xiang2017posecnn,manhardt2019explaining, park2022dprost}, while others utilize geometry-guided regression to determine the pose \cite{hu2020single, di2021so, wang2021gdr, liu2022gdrnpp_bop,feng2023nvr, zhang2022trans6d}. 
For RGBD image input, \cite{he2021ffb6d, he2020pvn3d, wang2019densefusion, jiang2022uni6d, mo2022es6d, zhou2023deep} integrate depth information with RGB images to estimate the object pose. 
Although instance-level methods have shown remarkable performance, they rely on the CAD model, necessitating costly data generation \cite{uy2021joint,di2023ured,zhang2024kpred,di2024shapematcher} or annotation. 
This dependency limits their practicality in real-world applications.

\subsection{Category-level Object Pose Estimation}
In the field of category-level object pose estimation, the objective is to determine the rotations, translations, and sizes of objects belonging to known categories. 
To define the pose of unseen objects, early research introduced the Normalized Object Coordinate Space (NOCS)~\cite{zhang2024lapose}, which defines a shared space for objects within a category. 
Subsequently, numerous studies based on RGBD data emerged, with some incorporating per-category shape prior information for pose estimation~\cite{spd,crnet,chen2021sgpa,zhang2022ssp,zhang2022rbp,li2023sd,fan2024acr,irshad2022shapo, irshad2022centersnap}, while others adopted a prior-free approach~\cite{chen2021fs, lin2021dualposenet, di2022gpv, goodwin2022zero, liu2023ist, lin2023vi, chen2024secondpose, lin2024ag, lunayach2024fsd,ikeda2024diffusionnocs}. 
Although these studies yield promising results, their reliance on depth information incurs higher costs and limits broader applications.

Recently, some works \cite{manhardt2020cpspp,lee2021category,fan2022object,wei2023rgb,zhang2024lapose} have attempted to estimate category-level object poses from a single RGB image to remove the dependence on depth. However, this is particularly challenging due to the inherent scale ambiguity. 
To overcome the challenges, MSOS~\cite{lee2021category} proposes a two-branch framework: one branch estimates the NOCS map, while the other reconstructs the metric-scale mesh to obtain the rendered depth. 
OLD-Net~\cite{fan2022object} introduces shape prior to enhance depth estimation. 
Both methods ultimately determine the final pose using the Umeyama~\cite{umeyama1991least} algorithm. 
DMSR~\cite{wei2023rgb} integrates the results of DPT models~\cite{ranftl2021dpt} and the shape prior, in addition to the RGB image, to estimate the NOCS map and metric scale, subsequently solving the pose utilizing the PnP algorithm~\cite{epnp2009ep}.
The aforementioned approaches all rely on classic indirect pose solvers, which might be suboptimal for not considering pose in optimizing neural networks.
Inspired by the success of the geometry-guided direct regression paradigm in the instance-level setting \cite{wang2021gdr,hodan2024bop,liu2022gdrnpp_bop}, more recently, LaPose~\cite{zhang2024lapose} extended the idea to the category-level task. 
Utilizing the Laplacian mixture model for NOCS map modeling and proposing scale-agnostic translation and size for end-to-end pose estimation, LaPose~\cite{zhang2024lapose} achieved superior performance compared to previous indirect approaches.
However, regressing category-level poses directly from instance-specific NOCS maps introduces redundant information and more overhead to the pose regression head. 
In this work, we further gradually eliminate the intra-class variation in the geometric representations to discard redundant information and achieve more efficient pose regression.

\subsection{Intermediate Representation for Object Pose Estimation}
Directly estimating the pose of an object from the input images is highly challenging. To address this issue, existing methods have proposed various intermediate representations to facilitate pose estimation.

In instance-level tasks, many early approaches \cite{rad2017bb8, peng2019pvnet, li2019cdpn, park2019pix2pose, wang2021gdr} first estimate dense or sparse 3D coordinate maps to establish 2D-3D correspondences before estimating the pose, which can be regarded as the most straightforward intermediate representation. 
EPOS \cite{hodan2020epos} introduces surface fragments to manage object symmetry more effectively. 
Zebra-Pose \cite{su2022zebrapose} encodes object surfaces via hierarchical binary grouping to address occlusion. 
NVR-Net \cite{feng2023nvr} employs two sets of 3D normal vectors to decouple rotation from translation.

In category-level tasks, NOCS \cite{wang2019normalized} adapts coordinate maps from instance-level tasks and further develops them into canonical object coordinate maps to satisfy the feasibility requirements of category-level tasks. 
SOCS \cite{wan2023socs} argues that a more effective representation for category-level tasks should incorporate semantic information, leading to the proposal of the Semantically-aware Object Coordinate Space and its corresponding coordinate map representation, known as the SOCS map. 
LaPose \cite{zhang2024lapose} addresses the shape uncertainty in RGB-based methods and introduces a NOCS map representation modeled using Laplacian distributions.

\begin{figure*}
\centering
\includegraphics[width=0.9\linewidth]{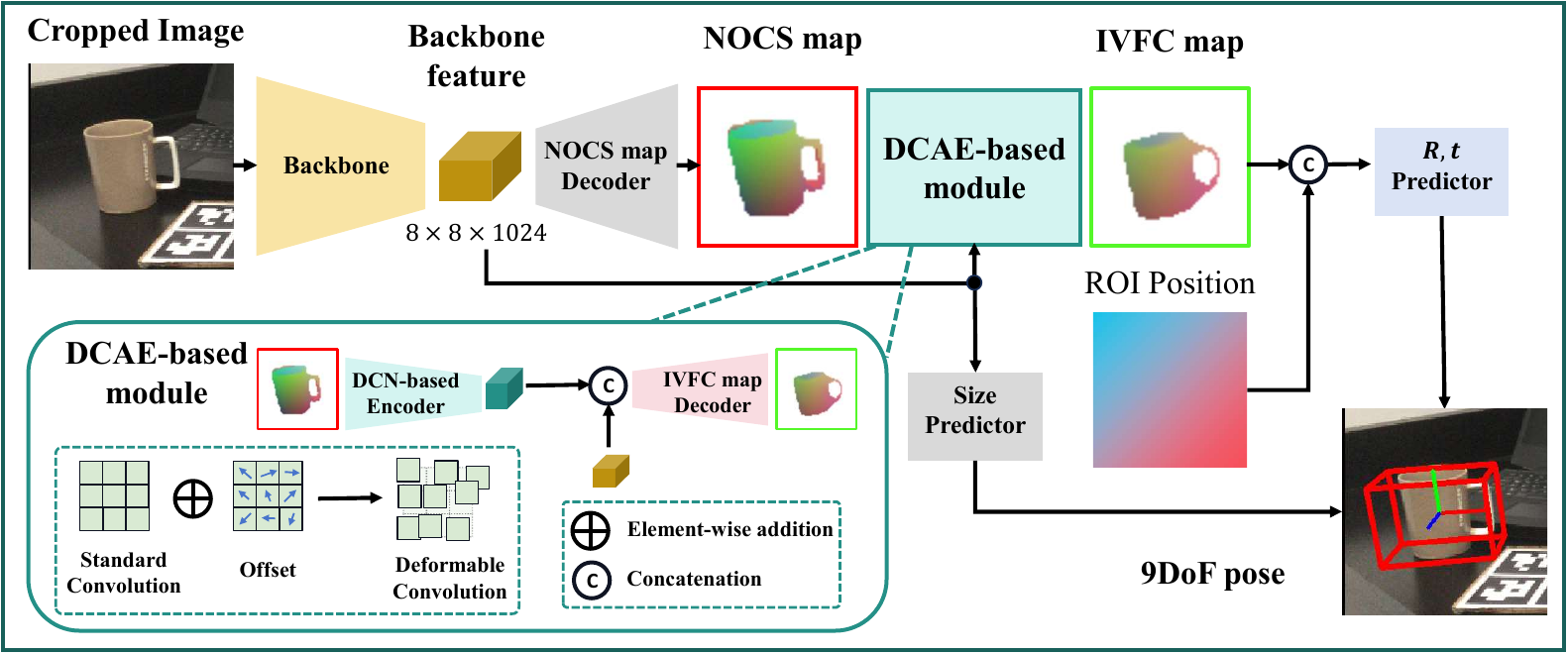}
\caption{\label{fig:pipline}
   Overview of our proposed GIVEPose. The core of our framework lies in the DCAE-based module, which facilitates the bridging of the NOCS map and our proposed IVFC map. Leveraging the NOCS map estimated from the backbone features, we employ a deformable convolutional encoder to selectively distill the information, thereby enabling the reconstruction of the IVFC map. This process gradually eliminates intra-class variations, ultimately yielding a more robust representation. By fusing the estimated IVFC map with the 2D positional information of the Region of Interest (ROI), we employ a lightweight rotation and translation (\(R,t\)) predictor. Concurrently, the object size is directly inferred from the backbone features.
}
\end{figure*}

\section{Method}
\subsection{Overview}
Given an RGB image with known camera intrinsics $K\in\mathbb{R}^{3\times 3}$, our objective is to estimate the 9DoF pose (3D rotation $R \in SO(3)$, 3D translation $t \in \mathbb{R}^3$, and 3D size $s\in \mathbb{R}^3$)  of the objects belonging to predefined categories within the image. 
We first follow \cite{lee2021category,fan2022object,wei2023rgb,zhang2024lapose} to obtain cropped images of the objects of interest using an off-the-shelf detector. 
Then we estimate the object pose from the cropped image by implementing a modified geometry-guided pose regression framework (as presented in \cref{fig:pipline}). This adaptation is informed by our proposed strategy of gradual intra-class variation elimination. 
Following the approach of LaPose~\cite{zhang2024lapose}, we estimate object size directly from backbone features. 
For clarity, it should be noted that throughout this paper, unless explicitly stated as 9DoF pose, the term ``pose'' refers exclusively to rotation and translation $P=[R|t]$, omitting size.

Specifically, we first present the intra-class variation problem when employing original geometry-guided monocular pose regression methods in category-level tasks (\cref{sec:rethinking}). Subsequently, we introduce the Intra-class Variation-Free Consensus (IVFC) map to tackle this problem and analyze the properties of both the NOCS map and the IVFC map (\cref{sec:inermediatemap}). Taking these properties and the nature of the coordinate map into account, we design a Denoising Autoencoder (DAE)-like network architecture based on deformable convolution to gradually eliminate intra-class variation (\cref{sec:reducevariation}). Finally, we summarize the loss function of this method (\cref{sec:loss}).

\subsection{Rethinking monocular category-level pose regression}
\label{sec:rethinking}
The geometry-guided monocular pose regression method has demonstrated strong performance in instance-level tasks \cite{hu2020single, wang2021gdr, di2021so, zhang2022trans6d, feng2023nvr}. Recently, LaPose~\cite{zhang2024lapose} adapted this approach for category-level tasks by modeling NOCS maps with the Laplacian Mixture Model, achieving state-of-the-art performance. 
However, we argue that the NOCS map is not an adequate intermediate representation for category-level tasks due to its intra-class variation problem.

\begin{figure}[t]
  \centering
   \includegraphics[width=0.92\linewidth]{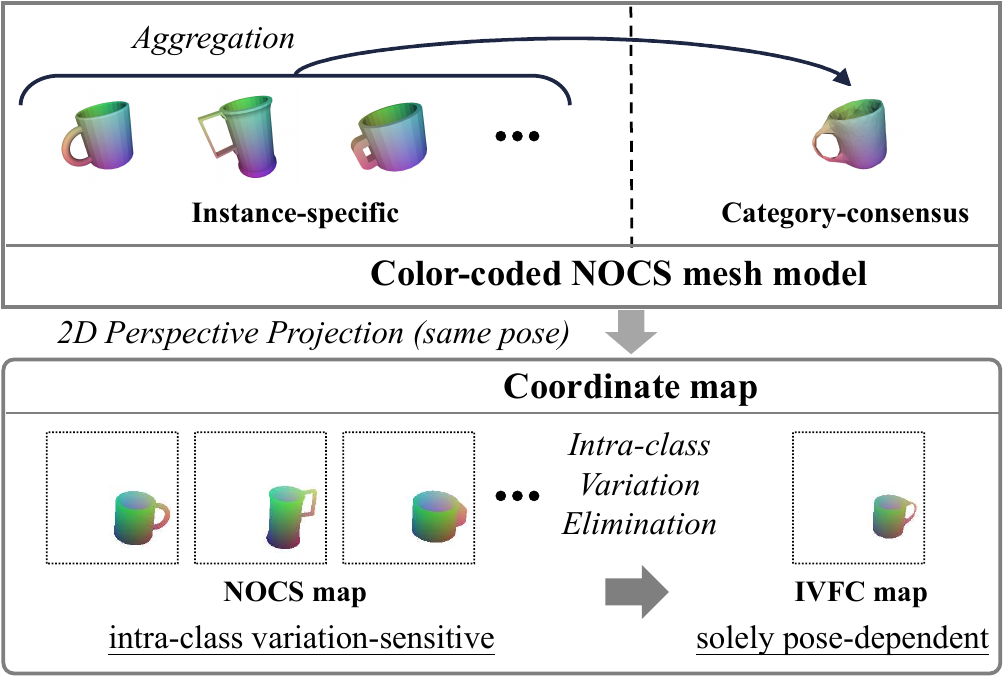}
   \caption{
   Illustration of the relationship between the NOCS and IVFC maps. Both coordinate maps are generated from color-coded NOCS models via perspective projection. The key difference lies in their origins: the NOCS map is derived from instance-specific models, whereas the IVFC map is derived from a category-consensus model. 
   The IVFC map represents a category-level shared coordinate system corresponding to category-level pose. Transforming NOCS maps to the IVFC map under the same pose eliminates intra-class variations.
   % We consider the IVFC map as a category-level shared coordinate map corresponding to the category-level pose. Consequently, the process of transforming different NOCS maps to the IVFC map under the same pose can be viewed as intra-class variation elimination.
   % \TODO{instead of many-to-one, one-to-one, we can say NOCS has redundant information for pose regression due to intra-class variation}
   }
   \label{fig:iv}
\end{figure}

\paragraph{Intra-class Variation Problem} 
To clarify the intra-class variation problem of the NOCS map, we first review how the NOCS map is generated. 
In NOCS~\cite{wang2019normalized}, different instances of the same class are aligned and normalized through the canonical pose to obtain models in the Normalized Object Coordinate Space. 
Given specific camera intrinsics and pose, the color-coded NOCS model's 2D perspective projection, i.e. the NOCS map, is obtained through rendering, illustrated in \cref{fig:iv} (left). This process can be described by the mapping $f_N$:
\begin{equation}
  f_N:(M_{NOCS},K,P)\to N_{map} ,
  \label{eq:fi}
\end{equation}
where $M_{NOCS}$ and $N_{map}$ respectively represent the color-coded NOCS model and the NOCS map.

Considering the scenario with fixed camera intrinsics, the NOCS map for a single object is solely related to its pose. 
In category-level tasks, however, the $M_{NOCS}$ of different instances within the same category exhibit intra-class shape variations. 
Consequently, the $f$-mapping of objects within the same category yields a NOCS map that contains intra-class variations. 
In instance-level tasks, we estimate the instance-level pose from the 2D-3D correspondences of the same instance. 
Whereas in category-level tasks, we must estimate the category-level pose from the 2D-3D correspondences of different instances belonging to the same category. 
This introduces the intra-class variation problem: the NOCS map contains redundant instance-specific detail information, so that the regression from NOCS-map-based 2D-3D correspondences to category-level pose becomes a more complicated many-to-one problem. This complexity increases the difficulty of network regression, making accurate pose estimation challenging.

% Illustration of the difference between the NOCS map (left) and the IVFC map (right). The left side displays the NOCS maps of various instances within the same category (mug) under identical pose. This demonstrates that, despite having similar contours and color distributions, the NOCS maps exhibit variability due to the intra-class shape variation of the instance-specific NOCS mesh models. In contrast, the IVFC map on the right, generated from the category consensus model, establishes a one-to-one mapping with the category-level pose by being invariant to redundant instance-specific information.\TODO{adapt caption according to new content}

\subsection{Intra-class Variation-free Consensus Map}
\label{sec:inermediatemap}
To address the intra-class variation problem, we introduce \textbf{I}ntra-class \textbf{V}ariation-\textbf{F}ree \textbf{C}onsensus map (IVFC map $C_{map}$). Specifically, rather than employing NOCS model for each instance during the projection process, we utilize a fixed public category-consensus model, denoted as $M_{Con}$. This model allows us to obtain category-level coordinate map (\cf \cref{fig:iv}), facilitating category-level pose estimation. The new mapping $f_C$ can be expressed as follows:
\begin{equation}
  f_C:(M_{Con},K,P)\to C_{map} .
  \label{eq:fc}
\end{equation}
% where $C_{map}$ denotes the IVFC map.

\begin{figure}[t]
  \centering
   \includegraphics[width=0.92\linewidth]{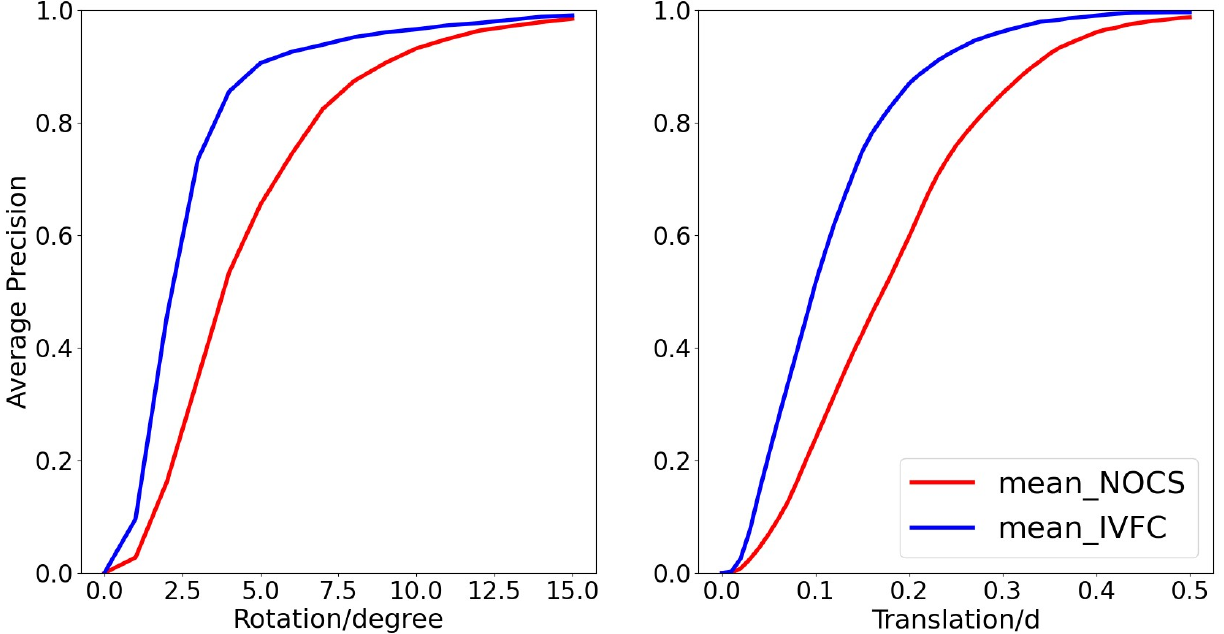}

   \caption{Comparison of mean average precision under various thresholds for scale-agnostic rotation and translation in direct pose estimation from different ground-truth coordinate maps: NOCS map (\textcolor{red}{red}) vs. IVFC map (\textcolor{blue}{blue}).}
   \label{fig:map_aps}
\end{figure}

% For the purpose of investigating the properties of two maps, we conducted comparative experiments utilizing two distinct coordinate maps independently for category-level pose regression. 
We showcase the distinction between the NOCS and IVFC maps with a comparative experimental analysis, independently utilizing each GT map for category-level pose regression.
Our findings, which are consistent with our theoretical analysis, indicate that learning to predict poses from the IVFC-map-based 2D-3D correspondences is indeed easier than from the NOCS-map-based 2D-3D correspondences, as illustrated in \cref{fig:map_aps}. 
However, when introducing intermediate geometric representations in regression tasks, the final results are influenced not only by the correspondence between the geometric representation and regression target but also by the difficulty of predicting the intermediate representation itself. 
In this context, the NOCS map is inherently easier to predict than our IVFC map due to its pixel-wise correspondence with input images, while the IVFC map lacks such pixel-wise alignment. 
Our experiments using the NOCS map and the IVFC map independently as intermediate representations yielded similar pose estimation accuracy. This suggests the redundancy effects in the NOCS map and the non-pixel-wise correspondence in the IVFC map both hinder category-level pose regression.
% However, when we directly replaced the NOCS map supervision with the IVFC map in pose regression framework, we did not observe the anticipated improvement in the final results' performance. Instead, the performances utilizing the two different maps separately were similar. We argue that while the IVFC map is superior to the NOCS map in estimating category-level pose in theory, we overlooked the fact that different maps present varying levels of prediction difficulty. Specifically, the NOCS map, which maintains a pixel-by-pixel correspondence with the original image, is easier to be estimated than the IVFC map which lacks such correspondence.

Let $D[p]$ represent the difficulty of fitting the conditional probability $p$, with larger values indicating greater difficulty, and $x$ denote the input image. 
The concept can be formally explained using the conditional probability formula as:
\begin{equation}
  \left\{\begin{matrix}
  D[p(P|N_{map})p(N_{map}|x)]\approx D[p(P|C_{map})p(C_{map}|x)] 
  \\D[p(N_{map}|x)]<D[p(C_{map}|x)]
  \\D[p(P|N_{map})]>D[p(P|C_{map})] 
  \end{matrix}\right. .
  \label{eq:prop}
\end{equation}

Based on the above analyses, we observe that although the final accuracies achieved using the two maps as intermediate representations are similar, the primary bottlenecks limiting their accuracies differ. 
When utilizing the NOCS map to supervise the intermediate layers, the intra-class variation problem complicates pose estimation. 
Conversely, when employing the IVFC map as supervision for the intermediate layers, the lack of pixel-by-pixel correspondence, as seen in the NOCS map, makes the prediction of coordinate maps more challenging.

\subsection{Gradual Intra-class Variation Elimination}
\label{sec:reducevariation}
 Considering the discussion above, we propose GIVEPose for category-level object pose estimation, which simultaneously leverages two complementary types of maps, capitalizing on their respective advantages for \textbf{G}radual \textbf{I}ntra-class \textbf{V}ariation \textbf{E}limination. 
 As illustrated in \cref{fig:pipline}, we first predict the NOCS map, which maintains a pixel-wise correspondence with the input image. 
 Subsequently, we integrate the feature encoded from the NOCS map and the backbone feature to estimate the IVFC map, which has a one-to-one mapping relationship with category-level object pose. 
 This process guides the network in progressively reducing intra-class variation, thereby enhancing the accuracy of pose estimation.

In order to bridge the gap between the NOCS map and our proposed IVFC map within the GIVEPose framework, we have developed a denoising autoencoder~\cite{vincent2008DAE} (DAE)-like architecture based on deformable convolution~\cite{wang2023dcnv3}. 
The original DAE can be utilized for image denoising, as described by the following equation:
\begin{equation}
    \tilde{x} =\Phi_{de}(\Phi_{en}(x')) , 
    \label{eq:ae}
\end{equation}
where $x'$ represents the noisy image, $\Phi_{en}$ denotes the encoder network, and $\Phi_{de}$ denotes the decoder network.

We employ a similar approach to eliminate intra-class variation. 
Specifically, we first predict the NOCS map based on the input image, and then we incorporate an approximate autoencoder structure. 
As depicted in \cref{fig:pipline}, after combining the features extracted by the encoder from the NOCS map with those from the backbone, we utilize a decoder to recover the IVFC map:
\begin{equation}
C_{map} =\Phi_{de}(\Phi_{en}(N_{map}),F_{image}),
    \label{eq:dae}
\end{equation}
where $F_{image}$ represents the backbone features extracted from the input cropped image. 
Our key insight is that each instance contains both category-consensus information and instance-specific details, which also applies to the NOCS map generated by the instance model. 
Using this DAE-like network, we eliminate instance-specific details while preserving the category-consensus information in the IVFC map, which is essential for category-level pose regression.

\paragraph{Deformable Convolutional Auto-Encoder (DCAE)} 
Most of the research utilizing Auto-Encoder (AE) in image processing follows the methodology in~\cite{masci2011CAE}, which introduces the Convolutional Auto-Encoder (CAE). 
However, eliminating intra-class variation is more challenging than traditional image denoising. 
The intra-class variation in the NOCS map is not superimposed on each pixel independently like ordinary noise, but corresponds to complex shape deformations in 3D space. 
Thus, a standard CAE is inadequate for this problem. 
Considering the two characteristics of this process: (1) the NOCS map and the IVFC map share similar contours but do not align pixel-by-pixel, and (2) the values in the coordinate map are correlated with pixel positions, we employ a Deformable Convolutional Auto-Encoder (DCAE) to achieve more robust feature extraction and more accurate IVFC map reconstruction as shown in~\cref{eq:dcae}:
\begin{equation}
C_{map} =\Phi_{de}(\Phi_{dcen}(N_{map}),F_{image}),
    \label{eq:dcae}
\end{equation}
where $\Phi_{dcen}$ is an encoder comprising deformable convolutions.

\subsection{Overall Training Loss}
\label{sec:loss}
The overall training loss function is as follows:
\begin{equation}
L = L_{pose}+\alpha L_{nocs}+\beta L_{ivfc},
 \label{eq:loss}
\end{equation}
where $L_{pose}$ is similar to LaPose~\cite{zhang2024lapose} for supervising the 9DoF pose learning, $L_{nocs}$ and $L_{ivfc}$ are used to supervise the learning of the NOCS map and the IVFC map, respectively (see Sup. Mat. for details). 
$\alpha$ and $\beta$ are hyperparameters used to balance the correspondence terms.

\begin{table*}
    \centering
    \tablestyle{16pt}{1.0}
    \begin{tabular}{@{}l|c c c|c c c c c@{}}
    % \toprule
        Method & $NIoU_{25}$ & $NIoU_{50}$ & $NIoU_{75}$ & $10^\circ 0.2d$ & $10^\circ 0.5d$ & $0.2d$ & $0.5d$ & $10^\circ$ \\ 
        \shline
        MSOS \cite{lee2021category} & 36.9 & 9.7 & 0.7 & 3.3 & 15.3 & 10.6 & 50.8 & 17.0 \\ 
        OLD-Net \cite{fan2022object} & 31.5 & 6.2 & 0.1 & 2.8 & 12.2 & 9.0 & 44.0 & 14.8 \\
        DMSR \cite{wei2023rgb} & 57.2 & 38.4 & 9.7 & 26.0 & 44.9 & 35.8 & 67.2 & 36.9 \\ 
        LaPose \cite{zhang2024lapose} & 70.7 & 47.9 & 15.8 & 37.4 & 57.4 & 46.9 & 78.8 & 60.7 \\ 
        \hline
        {Ours} &\textbf{71.4} & \textbf{50.3}& \textbf{20.8} & \textbf{44.6} & \textbf{64.8} & \textbf{48.9} & \textbf{81.1} & \textbf{67.8} \\ 
        % \bottomrule
    \end{tabular}
    \caption{\label{tab:norm}Quantitative comparison with existing methods on the REAL275 dataset using scale-agnostic evaluation metrics.}    
\end{table*}
\begin{table*}[t]
    \centering
    \tablestyle{16pt}{1.0}
    \begin{tabular}{@{}l|c c c|c c c c c@{}}
    % \toprule
        Method & $NIoU_{25}$ & $NIoU_{50}$ & $NIoU_{75}$ & $10^\circ 0.2d$ & $10^\circ 0.5d$ & $0.2d$ & $0.5d$ & $10^\circ$ \\ 
        \shline
        MSOS \cite{lee2021category} & 35.1 & 9.9 & 0.8 & 5.9 & 31.6 & 48.6 & 8.9 & 47.2 \\ 
        OLD-Net \cite{fan2022object} & 50.4 & 14.5 & 0.5 & 11.3 & 39.9 & 53.3 & 17.2 & 60.5 \\
        DMSR \cite{wei2023rgb} & 74.4 & 46.0 & 11.1 & 38.6 & 68.1 & 74.4 & 42.8 & 79.9 \\ 
        LaPose \cite{zhang2024lapose} & 75.3 & 49.4 & 14.1 & 42.4 & 73.1 & 80.0 & 45.4 & 81.2\\ 
        \hline
        {Ours} &\textbf{76.1} & \textbf{53.3}& \textbf{19.4} & \textbf{47.5} & \textbf{75.5} & \textbf{82.4} & \textbf{49.7} & \textbf{81.7} \\ 
        % \bottomrule
    \end{tabular}
    \caption{Quantitative comparison with existing methods on the CAMERA25 dataset \cite{wang2019normalized} using scale-agnostic evaluation metrics.}
    \label{tab:C_norm}
\end{table*}
\section{Experiments}
\subsection{Experimental Setup}
\paragraph{Datasets}
To verify the effectiveness of our proposed GIVEPose, we conducted comprehensive experiments on three datasets: CAMERA25, REAL275, and Wild6D~\cite{fu2022wild6d}. 

The CAMERA25 and REAL275 datasets, introduced by NOCS~\cite{wang2019normalized}, have been widely used in previous studies and serve as standard benchmarks for evaluation. CAMERA25 consists of synthetic data derived from ShapeNetCore~\cite{chang2015shapenet}, encompassing 1,269 object models across six categories (bottle, bowl, camera, can, laptop, and mug). 
Complementing this, REAL275 provides real-world scenarios, featuring 42 physical object instances from the same six categories, captured across 18 scenes using a structured sensor. From these scenes, 2,750 images from 6 designated scenes are allocated for testing purposes. 

To further validate the generalization capability of our method, we extended our evaluation to the Wild6D~\cite{fu2022wild6d} dataset, which comprises 5,166 videos containing 1,722 object instances across five categories (bottle, bowl, camera, laptop, and mug). For evaluation purposes, 486 videos were designated as the test set in Wild6D~\cite{fu2022wild6d}.

\paragraph{Evaluation Metrics}
To exclude the effects of scale ambiguity, we follow LaPose~\cite{zhang2024lapose} by utilizing scale-agnostic metrics for our evaluation. 
Specifically, we report the mean Average Precision (mAP) of Normalized 3D Intersection over Union with threshold of $\xi\,\%$ ($NIOU_{\xi}$) and the mAP of $n^\circ m d$, which indicates pose precision with rotation error less than $n^\circ$ and normalized translation error less than $m$ times the diagonal length $d$ of the tight object bounding box.
For a more comprehensive comparison with previous methods, we also present results using evaluation metrics that incorporate absolute object scale as a reference.

\begin{table}
    \centering
    \tablestyle{10pt}{1.0}
    \begin{tabular}{@{}l|c c|c c@{}}
    % \toprule
        Method & $IoU_{25}$ & $IoU_{50}$ & $10^\circ10cm$ & $10cm$\\ 
        \shline
        MSOS \cite{lee2021category} & 33.2 & 13.6 &  11.8 & 43.4\\ 
        OLD-Net \cite{fan2022object} & 26.4 & 7.7 &  8.6 & 31.4\\
        DMSR \cite{wei2023rgb} & 37.4 & 16.3 &  25.2 & 40.0\\ 
        LaPose \cite{zhang2024lapose} & 41.2 & 17.5 &  30.5&44.4\\ 
        \hline
        {Ours} &\textbf{42.9} & \textbf{20.1}&  \textbf{34.2} & \textbf{45.9}\\ 
        % \bottomrule
    \end{tabular}
    \caption{Quantitative results on the REAL275 dataset using evaluation metrics with absolute object scale.}
    \label{tab:abs}
\end{table}
\paragraph{Implementation Details}
As in previous research~\cite{fan2022object, wei2023rgb, zhang2024lapose}, we employ MaskRCNN~\cite{maskrcnn} to detect instances, utilizing the results provided by~\cite{lin2021dualposenet}. 
For the category-consensus model utilized to generate the IVFC map, we first obtained the per-category shape prior as described in SPD~\cite{spd}. 
Additionally, we employed the surface reconstruction technique available in Open3D to reconstruct the consensus mesh models (see Sup. Mat. for details). 
In our proposed DCAE-based module, we employ the deformable convolution proposed by~\cite{wang2023dcnv3}. 
In all ablation experiments, we trained each model for 300 epochs with a batch size of 48.

% \begin{figure}[t]
%   \centering
%    \includegraphics[width=1.0\linewidth]{fig/precise_compare.pdf}
%    \caption{Precise comparison of LaPose (top) and Ours (bottom) on scale-agnostic 3D IoU, rotation and translation on REAL275 dataset.}
%    \label{fig:precise}
% \end{figure}

\begin{table}
    \centering
    \tablestyle{4pt}{1.1}
    \begin{tabular}{@{}l|c c c|c c@{}}
    % \toprule
        Method & $NIoU_{25}$ & $NIoU_{50}$ & $NIoU_{75}$ & $10^\circ 0.2d$ & $10^\circ 0.5d$\\ 
        \shline
        LaPose \cite{zhang2024lapose} & 84.8 & 60.3 & 24.4 & 32.9 & 37.1 \\ 
        \hline
        {Ours} &\textbf{87.3} & \textbf{60.4}& \textbf{28.1} & \textbf{33.4} & \textbf{40.3} \\ 
        % \bottomrule
    \end{tabular}
    \caption{Quantitative comparison with LaPose~\cite{zhang2024lapose} on the Wild6D dataset using scale-agnostic evaluation metrics.}
    \label{tab:wild}
\end{table}
\subsection{Comparison with Existing Methods}
\nbf{Results on NOCS datasets} We compare our GIVEPose with four existing RGB-based category-level pose estimation methods~\cite{lee2021category, fan2022object, wei2023rgb, zhang2024lapose}. 
\cref{tab:norm} and \cref{tab:C_norm} present the absolute-scale metric results on the REAL275 and CAMERA25 datasets, respectively. 
It is evident that our method significantly outperforms the existing methods across all metrics in both datasets. 
Specifically, our method demonstrates superior performance on the REAL275 dataset, surpassing previous state-of-the-art method LaPose~\cite{zhang2024lapose} by 5.0\,\% and 7.2\,\% on the most stringent metrics, $NIoU_{75}$ and $10^\circ 0.2d$, respectively. 
Furthermore, on the synthetic CAMERA25 dataset, our approach also exhibits notable improvements over the leading benchmark with increases of 5.3\,\% and 5.1\,\% for $NIoU_{75}$ and $10^\circ 0.2d$, respectively. 

% \cref{fig:precise} presents a more detailed analysis. As illustrated in the graph, our method shows particularly significant improvements in categories with high intra-class variation (e.g., camera), which can be attributed to our gradual intra-class variation elimination strategy based on the IVFC map. For the bottle category, the performance of our method is slightly lower. We hypothesize that this may be due to our approach of training a single model across multiple categories, which could introduce some inter-category influences. Nevertheless, our method continues to outperform others in terms of overall average results.

For a comprehensive comparison, following LaPose~\cite{zhang2024lapose}, we utilize an independent network to estimate object scales and compute absolute scale metrics on the REAL275 dataset. 
The results are presented in \cref{tab:abs}, which also demonstrate the superiority of GIVEPose. 

\begin{figure*}
  \centering
   \includegraphics[width=0.95\linewidth]{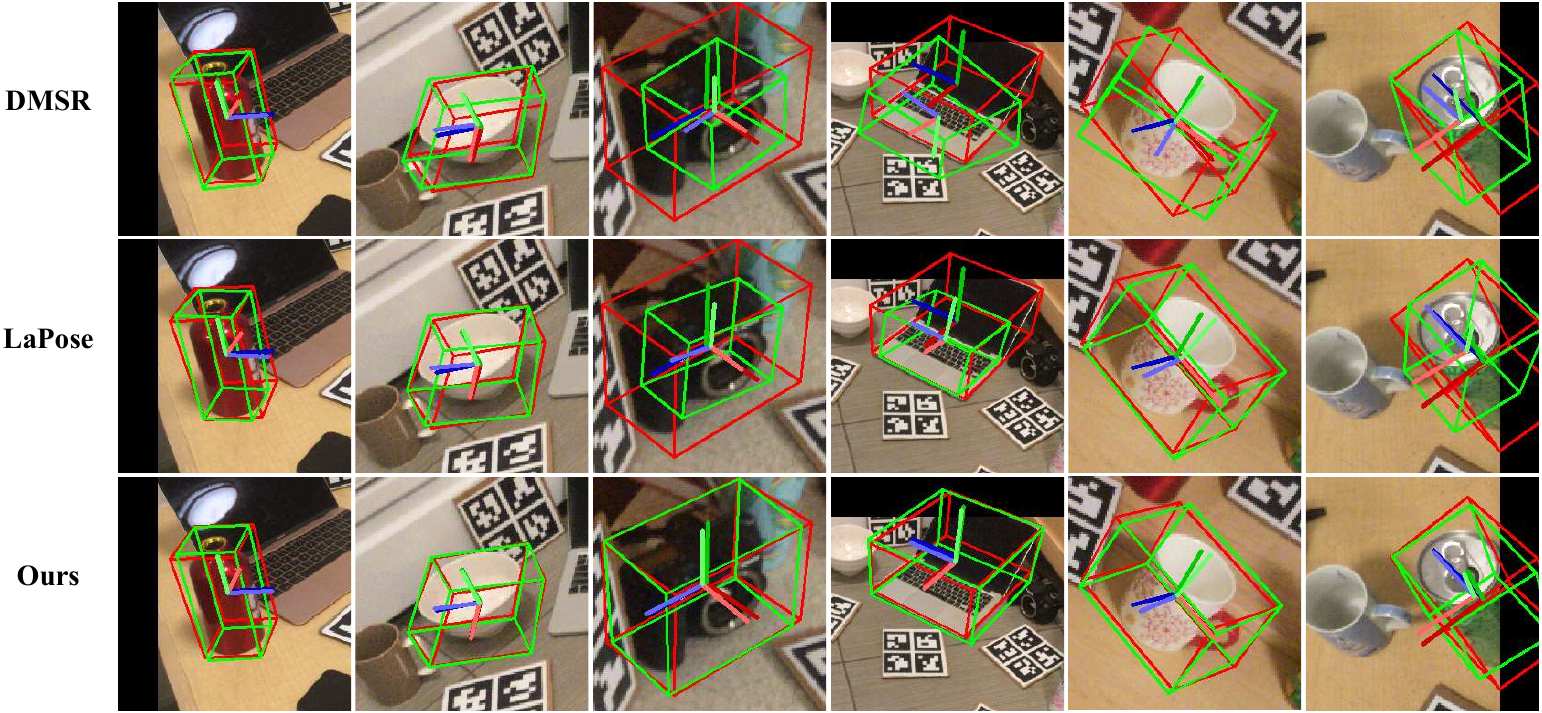}

   \caption{Qualitative comparison with LaPose~\cite{zhang2024lapose} and DMSR~\cite{wei2023rgb} on REAL275. \textcolor{red}{Red} and \textcolor{green}{green} boxes denote the GT and predicted results. For the axis projections, darker shades indicate the ground truth, while lighter shades correspond to the predicted results.}
   \label{fig:visual}
\end{figure*}

Furthermore, \cref{fig:visual} presents a qualitative comparison of our method with the top two existing methods. 
Our GIVEPose generally outperforms the existing methods, particularly when handling categories with large intra-class variations (e.g., camera). 
This is attributed to the gradual elimination of intra-class variation with the DCAE-based module. 
Moreover, our method exhibits improved robustness when dealing with truncated objects.

\paragraph{Results on the Wild6D dataset}
To validate the generalization capability of our model, following IST-Net~\cite{liu2023ist}, we directly evaluated the model trained on CAMERA25 and REAL275 datasets on the larger-scale Wild6D test set, comparing it with the current state-of-the-art method LaPose~\cite{zhang2024lapose}. 
As shown in \cref{tab:wild}, our method achieves notable improvements against LaPose on Wild6D, further demonstrating its robustness and generalization ability.

\begin{table*}
    \centering
    \tablestyle{16pt}{1.0}
    \begin{tabular}{@{}l|l|c c|c c@{}}
    % \toprule
        Row & Method & $NIoU_{50}$ & $NIoU_{75}$ & $10^\circ 0.2d$ & $10^\circ 0.5d$  \\ 
        \shline
        A0 & GIVEPose (Ours) & 50.3 & \textbf{20.8} & \textbf{44.6} & \textbf{64.8}  \\ 
        \hline
        B0 & w/o NOCS map supervision   & 47.9 & 18.2 & 41.4 & 60.1 \\
        B1 & w/o IVFC map supervision  & 46.8 & 16.4 & 41.2 & 63.8 \\ 
        B2 & directly concat& \textbf{50.5} & 18.0 & 40.7 & 59.7 \\
        \hline
        C0 & DCAE $\to$ CAE   & 49.7 & 17.5 & 40.5 & 59.0 \\ 
        C1 & DCAE $\to$ ViT-AE& 49.0 & 18.3 & 41.3 & 61.8 \\
        \hline
        D0 & IVFC map $\to$ Ball map& 48.0 & 17.8 & 41.1 & 64.0 \\

        % \bottomrule
    \end{tabular}
    \caption{Ablation study on the REAL275 dataset. All ablations are conducted w.r.t. A0.}
    \label{tab:ablation}
\end{table*}
\subsection{Ablation Studies}
\label{sec:ablation}
To validate the effectiveness of our design choices, we performed a series of ablation studies on the REAL275 dataset, with the results presented in \cref{tab:ablation}.

\paragraph{Efficacy of Gradual Intra-class Variation Elimination}
To assess the efficacy of our proposed Gradual Intra-class Variation Elimination strategy, we conducted ablation studies by independently removing the supervision of the NOCS map and the IVFC map while maintaining a constant number of network parameters. 
The results demonstrate that the absence of either map leads to a substantial decrease in network performance (B0)-(B1). 
In particular, the absence of NOCS map supervision led to performance reductions from 44.6\,\% to 41.4\,\% and from 64.8\,\% to 60.1\,\% for the $10^\circ 0.2d$ and $10^\circ 0.5d$ metrics, respectively (B0). 
Moreover, eliminating the supervision of the IVFC map caused a 4.4\,\% decline in NIOU75 (B1).
Additionally, we attempted to predict both maps from the backbone feature in parallel, concatenating the two estimated maps and inputting them into the final pose predictor along with 2D positions (B2). 
Experimental evidence indicates that this rudimentary approach fails to effectively integrate advantages of both maps and yields inferior results compared to the GIVE strategy. 

\paragraph{Efficacy of DCAE}
% In the process of predicting IVFC maps from NOCS maps, we propose the implementation of a DCAE-based module. 
We proposed incorporating a DCAE-based module to enhance the prediction of IVFC maps from NOCS maps.
To evaluate its indispensability, we conducted comparative experiments by replacing the deformable convolution of the DCAE-based module with standard convolution \cite{lecun1998gradient} and Vision Transformer (ViT) \cite{dosovitskiy2020image}, respectively.
The results in \cref{tab:ablation} (C0)-(C1) demonstrate a significant performance degradation when standard convolution is employed in place of deformable convolution. 
Specifically, we observed a 4.1\,\% decrease on the $10^\circ 0.2d$ metric and a 5.8\,\% decrease on the $10^\circ 0.5d$ metric (C0). 
While the utilization of ViT yields marginally superior results compared to standard convolution, it still underperforms relative to deformable convolution by 3.5\,\% on the $10^\circ 0.2d$ metric and 2.5\,\% on the $NIoU_{75}$ metric (C1).
These findings underscore the critical role of our proposed DCAE-based module in enhancing the overall performance.

\nbf{Efficacy of the IVFC map}
% In \cref{sec:inermediatemap}, we mentioned that 
If only a one-to-one correspondence between the coordinate map and pose is required, the category-consensus 3D model can be arbitrarily fixed in the normalized object coordinate space, such as a simple sphere. 
We designate the coordinate map generated by projecting the sphere as the ``Ball map''. 
To evaluate the efficacy of our proposed IVFC map, we conducted ablation by replacing the IVFC map with the Ball map.
While using the Ball map has already demonstrated strong performance, the results show that employing the IVFC map leads to further improvements across all metrics (D0). 
We hypothesize that this enhancement stems from the IVFC map's inherent ability to capture shared information across instances within a category. 
While lacking pixel-wise correspondence with the NOCS map, the IVFC map preserves similar contours and value distributions with the NOCS map,
making it well-suited for category-level pose estimation.

\section{Conclusion, Limitation and Future Work}
This paper has introduced GIVEPose, a novel approach for addressing intra-class shape variation in RGB-based category-level object pose estimation via gradual intra-class variation elimination. 
Our method overcomes the inherent limitations of relying solely on the NOCS map in category-level object pose estimation tasks, which leads to intra-class variation problem. 
To mitigate this problem, we propose the IVFC map as a complementary representation.
Leveraging the characteristics of both NOCS and IVFC maps, we have developed a framework that employs a deformable convolution-based module to facilitate the gradual elimination of intra-class variation. 
By eliminating redundant instance-specific information, we can estimate category-level object pose more accurately.
Our extensive experiments conducted on the NOCS dataset demonstrate that GIVEPose achieves significant performance improvements over existing state-of-the-art approaches.

\nbf{Limitation and Future Work}
Our training process relies on annotated real-world data, which is expensive and time-consuming to acquire. 
Additionally, since our method operates independently of the detection network, its overall accuracy is inherently constrained by the detector's performance. 
Potential solutions such as self-supervised learning strategies \cite{lin2022category,wang2021occlusion} and unified architectures \cite{Misra_2021_ICCV} could be explored in future work to address these limitations.

\smallskip
\nbf{Acknowledgements.} This work was supported by the National Key R\&D Program of China under Grant 2018AAA0102801, the National Natural Science Foundation of China under Grant No.~62406169, the China Postdoctoral Science Foundation under Grant No.~2024M761673, and the Shenzhen Key Laboratory of Next Generation Interactive Media Innovative Technology (No.~ZDSYS20210623092001004).

% This work was supported in part by the National Natural Science Foundation of China under Grant No.~62406169, 
% and in part by the China Postdoctoral Science Foundation under Grant No.~2024M761673. 
{
    \small
    \bibliographystyle{ieeenat_fullname}
    \bibliography{main}
}


\end{document}